%% file: root.tex
\documentclass[letterpaper, 10 pt, conference]{ieeeconf}      % Use this line for a4 paper
\IEEEoverridecommandlockouts                              % This command is only needed if 
\usepackage{float}
\usepackage{graphics}
\usepackage{multirow}
\usepackage{amsmath}
\usepackage{amsmath,amssymb}
\DeclareMathOperator*{\argmax}{arg\,max}

\usepackage{algorithm,algorithmicx}
\usepackage{algorithm} 
\usepackage[noend]{algpseudocode}

\makeatletter
\let\NAT@parse\undefined
\makeatother
\usepackage[numbers,sort&compress]{natbib}
\usepackage{hyperref}
\usepackage{graphicx}

\usepackage{xcolor}

\title{\LARGE \bf
RoboAssembly: Learning Generalizable Furniture Assembly Policy\\ in a Novel Multi-robot Contact-rich Simulation Environment
}

\author{Mingxin Yu*$^{1}$, Lin Shao*$^{2}$, Zhehuan Chen$^{1}$, Tianhao Wu$^{1}$, Qingnan Fan$^{3}$, Kaichun Mo$^{2}$, and Hao Dong$^{1}$% <-this % stops a space
\thanks{*These authors contribute equally and share the first authorship.}% <-this % stops a space
\thanks{$^{1}$CFCS, Computer Science Department, Peking University, China.
        {\tt\small \{1700011374, 1900013017, hao.dong\}@pku.edu.cn}}%
\thanks{$^{2}$Artificial Intelligence Lab, Stanford University, USA.
        {\tt\small \{lins2, kaichunm\}@stanford.edu}}%
\thanks{$^{3}$Visual Computing Center of Tencent AI Lab, China.
        {\tt\small fqnchina@gmail.com}}%
}

\begin{document}

\maketitle
\thispagestyle{empty}
\pagestyle{empty}

%%%%%%%%%%%%%%%%%%%%%%%%%%%%%%%%%%%%%%%%%%%%%%%%%%%%%%%%%%%%%%%%%%%%%%%%%%%%%%%%
\begin{abstract}
\input{tex/abs}
\end{abstract}

%===============================================================================
\section{Introduction}
\input{tex/intro}\label{sec:intr}

\section{Related Work}
\input{tex/relatedwork}\label{sec:relatedwork}

\section{Simulation Environment and Assets}
\input{tex/datasim}\label{sec:datasim}

\section{Reinforcement Learning Formulation}
\input{tex/probdef}\label{sec:prof}

\section{Technical Approach}
\input{tex/approach}\label{sec:tech}

\section{Experiments}
\input{tex/exp}\label{sec:exp}

\section{Conclusion}
\label{sec:conclusion}
\input{tex/con}\label{sec:con}
%======================

{\small
\bibliographystyle{IEEEtranN}
\bibliography{references}
}

%\section{Supplementary}
%\label{sec:supp}
%\input{tex/supp}

\end{document}

%% file: tex/abs.tex
Part assembly is a typical but challenging task in robotics, where robots assemble a set of individual parts into a complete shape. In this paper, we develop a robotic assembly simulation environment for furniture assembly. We formulate the part assembly task as a concrete reinforcement learning problem and propose a pipeline for robots to learn to assemble a diverse set of chairs. Experiments show that when testing with unseen chairs, our approach achieves a success rate of 74.5\% under the object-centric setting and 50.0\% under the full setting. We adopt an RRT-Connect~\cite{844730} algorithm as the baseline, which only achieves a success rate of 18.8\% after a significantly longer computation time. Supplemental materials and videos are available on our project webpage:~\href{https://sites.google.com/view/roboticassembly}{https://sites.google.com/view/roboticassembly}.

%% file: tex/intro.tex
Endowing robots with the ability to autonomously assemble enormous products from their full sets of parts has the potential to profoundly change the manufacturing industry. Although the automation level has increased in the last decades, autonomous robotic assembly remains a challenging task. A typical robotic assembly pipeline consists of action sequences repeating the following stages: (1) picking up a particular part, (2) finding a feasible collision-free path to move it to an appropriate 6D pose, (3) mating it precisely with the other parts, and (4) placing the assembled parts to the ground and preparing for the next pickup movement. Robots need to know which part to choose for the attachment at each time and in what order to assemble all the parts into a functionally and physically valid shape. To connect two parts, robots not only need to master multiple contact-rich manipulation primitive skills such as grasping, placing, and object re-orientation, but also manage complex multi-arms or multi-robots cooperation.

\begin{figure}[thb!]
 \centering
 \includegraphics[width=\linewidth]{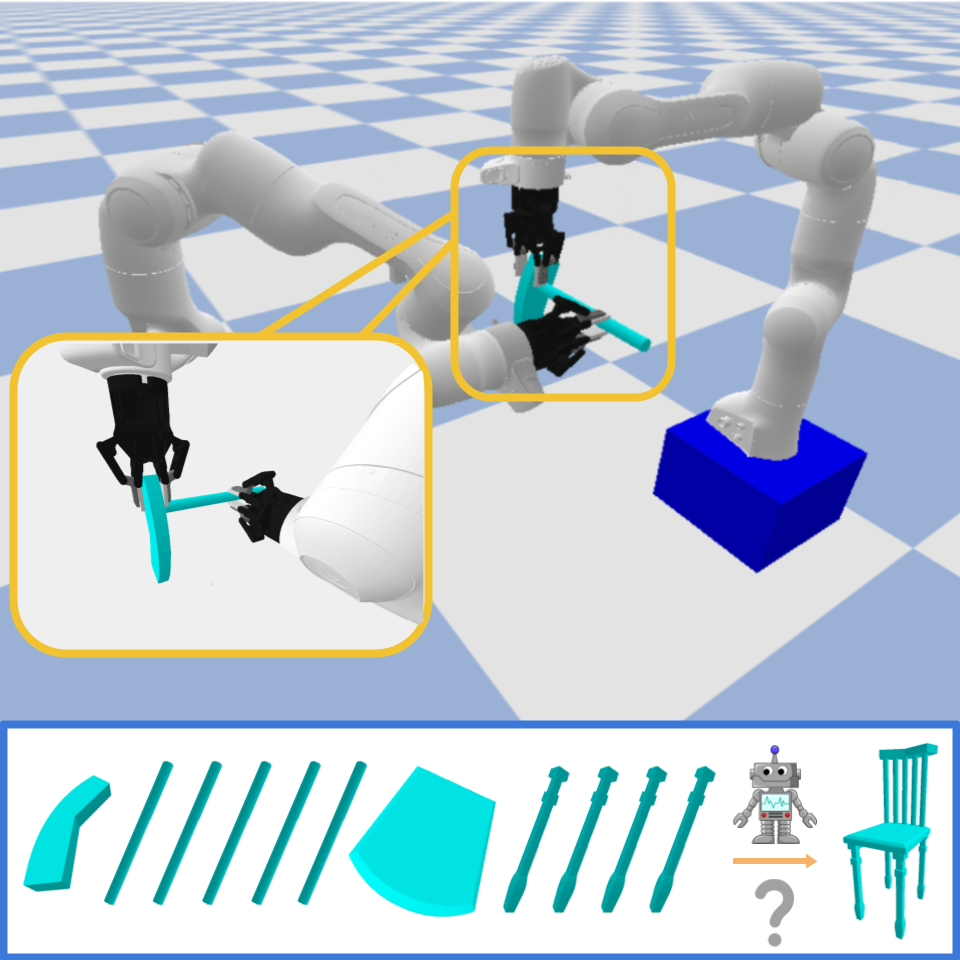}
 \caption{Visualization of our robotic assembly simulation environment. Based on our simulation environment, we formulate the assembly process as a reinforcement learning problem. Two Franka arms mounted on mobile platforms are assembling a chair cooperatively.}
\label{fig:teaser}
\vspace{-6mm}
\end{figure}

Many previous works in robotics and computer vision attempted to tackle this challenging problem from different aspects. One line of works ignores the physical interaction and focuses on estimating the desired poses~\cite{li2020learning, HuangZhan2020PartAssembly}. Other works teach robots to master complex object manipulation skills, such as object grasping~\cite{mahler2017dex,shao2020unigrasp}, object reorientation~\cite{cheng2021learning},peg-insertion~\cite{thomas2018learning,luo2019reinforcement}, tool manipulation~\cite{shao2020learning}. Some works formulate the assembly as a planning problem and propose various approaches~\cite{Halperin98, Lee1992BackwardAP, hartmann2021long}. ~\citet{suarez2018can} demonstrate that, based on state-of-the-art robotic capabilities, it is possible to physically assemble an IKEA chair with two robots, soliciting all manipulation skills. However, their sequence was hard-coded through considerable engineering efforts. Recently, \emph{the IKEA Furniture Assembly Environment}~\cite{lee2021ikea}, most similar to our work, simulates complex long-term robotic manipulation tasks with 80 furniture models using different robots based on Mujoco~\cite{6386109}. The environment leverages the weld equality constraint provided by the Mojuco~\cite{6386109} to merge two select parts as long as they are near to each other within a threshold. 

Our environment uses multi-robots to assemble furniture. Multi-robots greatly expand the working space compared to a single-arm robot with a fixed base and can perform more complex manipulation operations. Motion-planning modules are integrated into our simulation environments to enable the multi-robots to perform complicated manipulation operations. Our environment also considers strict collision constraints in robotic manipulation operations. We build a dataset with 220 chairs from the PartNet dataset~\cite{mo2019partnet}, annotated with additional connection points and graspable regions, as our simulation assets, to investigate robotic assembly over realistic and complicated shapes. Moreover, we formulate the assembly process as a concrete and feasible reinforcement learning problem. We combine a structured representation with a model-free RL algorithm that assembles a diverse set of chairs. We leverage the imitation learning process to train a unified policy network that can assemble various shapes. Experiments show that our approach achieves a 74.5\% success rate under an object-centric setting and a 50\% success rate under the full setting when testing unseen chairs. We adopt a motion planning algorithm called RRT-Connect~\cite{844730} which only achieves an 18.8\% with significantly more planning steps.

In summary, our contributions are: 1) we develop a robotic simulation environment for furniture assembly, 2) we formulate the furniture assembly process as a concrete and feasible reinforcement learning problem, 3) we propose a novel policy learning pipeline to assemble a diverse set of chairs with multi-robots, and 4) we demonstrate that the learned policy also generalizes to unseen novel shapes.

%% file: tex/relatedwork.tex
\subsection{Simulation Environments for Robotic Manipulation}
Most robot manipulation simulations often concentrate on solving specific manipulation problems, such as grasping~\cite{Miller2004, mahler2017binpicking} and in-hand manipulation~\cite{andrychowicz2020learning,rajeswaran2017learning,rollergrasperV2}. The Recent MetaWorld~\cite{yu2020meta} and RLBench~\cite{james2020rlbench} contain a variety of manipulation tasks for robot learning. But these tasks are all short-term object manipulation tasks. The work most related to ours is \emph{the IKEA Furniture Assembly Environment}~\cite{lee2021ikea}. It simulates complex long-term robotic manipulation tasks with 80 furniture models and different robots. Unlike the IKEA environment~\cite{lee2021ikea}, our simulation environment supports the multi-robots assembly setting. Multiple 3DoF mobile platforms mounted with 7-DoF robotic arms shown in Fig.\ref{fig:teaser} are designed. They significantly expand the working space during the assembly process. To enable them to navigate, manipulate, and co-operate to successfully assemble furniture, motion-planning modules are integrated into our simulation environments. Usually, motion planning modules take time to compute; we improve the speed so that we can run popular deep reinforcement learning algorithms within reasonable time periods. When merging two select parts, the IKEA furniture environment leverages the weld equality constraint provided by the Mojuco~\cite{6386109} to merge two parts as long as they are near to each other within a threshold. Our simulation environment considers more strict part attachment processes. Our environment simulates the precise collision-checking and rigid-body contact processes. It has realistic collision constraints in robotic manipulation operations such as pick-and-place and executing motion trajectories.

\subsection{Assembly Planning}
Assembly is a long-horizon manipulation task, and many works formulate it as a planning problem focusing on finding valid sequences to assemble a product from its parts. \citet{Natarajan88} shows that the complexity of deciding the existence of an assembly sequence is PSPACE-hard in general. \citet{KAVRAKI1995159} demonstrate that partitioning a planar assembly into two connected parts remains NP-complete. However, several assembly strategies have been investigated. \citet{Halperin98} presents a general framework for finding assembly motions based on the concept of motion space where each point represents a mating motion independent of the moving part set. 
\citet{Lee1992BackwardAP} develops a backward assembly planner to reduce the search space by merging and grouping parts based on interconnection feasibility.
More recently, \citet{hartmann2021long} integrate task and motion planning~(TAMP)~\cite{garrett2021integrated} in robotic assembly planning and propose a rearrangement pipeline for construction assembly. 
Different from these planning approaches, we formulate the assembly process as a reinforcement learning problem and train a model-free RL policy learning to assemble a diverse set of chairs and generalize to unseen chairs.

\subsection{Learning to Assemble}
Thanks to the recently proposed large-scale object-part datasets~\cite{mo2019partnet,yi2016scalable}, 
recent works~\cite{li2020learning, huang2020generative} in computer vision have investigated 6-DoF pose prediction for 3D shape part assembly. \citet{li2020learning} learns to assemble 3D shapes from a single image as guidance. \citet{huang2020generative} removes the external image guidance and learns a graph-based generative model. However, these works tackle the part assembly problem from the perception side; the robotic assembly task in the real world is much more complex. Robots need to decide how to pick up the parts and find feasible collision-free paths to mate two parts together, which is the main goal of this work.

%% file: tex/datasim.tex
Using bullet~\cite{coumans2021}, we build a robotic assembly simulation environment where two robots need to pick up two parts sequentially, plan collision-free motion trajectories, mate two parts precisely, and return the next pickup movement. We adopt 220 chairs from the PartNet dataset~\cite{mo2019partnet}, adjust the part relative poses to satisfy collision and contact conditions, and annotate connection points and graspable regions used as our simulation assets to study the robotic assembly task.

\subsection{Simulation Environment}
We design two settings which are the {\bf object-centric setting} and the {\bf full setting}. In the object-centric setting, no robots are loaded. We allow users to directly control each part to be moved to a pose specified by users. We describe what type of actions are provided to merge two parts in Sec.\ref{sec:step_fun}. After the merging process, the merged parts are positioned on the ground. We describe the penalty function for collision and incorrect merging action in the Sec.\ref{sec:step_fun}.

In the full setting, two robots are loaded in the simulation. Each of them is a seven DoF Franka-Arm mounted on a three DoF mobile platform. These two robots significantly improve the working space for the assembly process. The robots would pick up parts, merge them, and place the merged parts sequentially. There are several fixed-base holders in the environment to hold the parts when these parts are not grasped by the robots. We integrate the motion-planning modules in our simulation environment to control the two robots to navigate, cooperate, and conduct complicated manipulation operations. We integrate the OMPL library~\cite{sucan2012ompl} in our simulation environment and run the RRT-connect algorithm~\cite{844730} to search for a feasible path from the initial configuration to the goal configuration. To speed up the simulation process, we replace the pybullet default collision checking function with FCL~\cite{6225337} and re-implement the checking modules in the FCL to work for rigid articulation bodies. The simulation process is sped up to the degree that popular deep reinforcement learning algorithms can be applied in our simulation environment. We describe how to merge two parts and the corresponding reward function in Sec.\ref{sec:step_fun}. 

Due to the length limitation, videos illustrating the simulation environments can be found on our project webpage:~\href{https://sites.google.com/view/roboticassembly}{https://sites.google.com/view/roboticassembly}
\begin{figure}[t!]
 \centering
 \includegraphics[width=\linewidth]{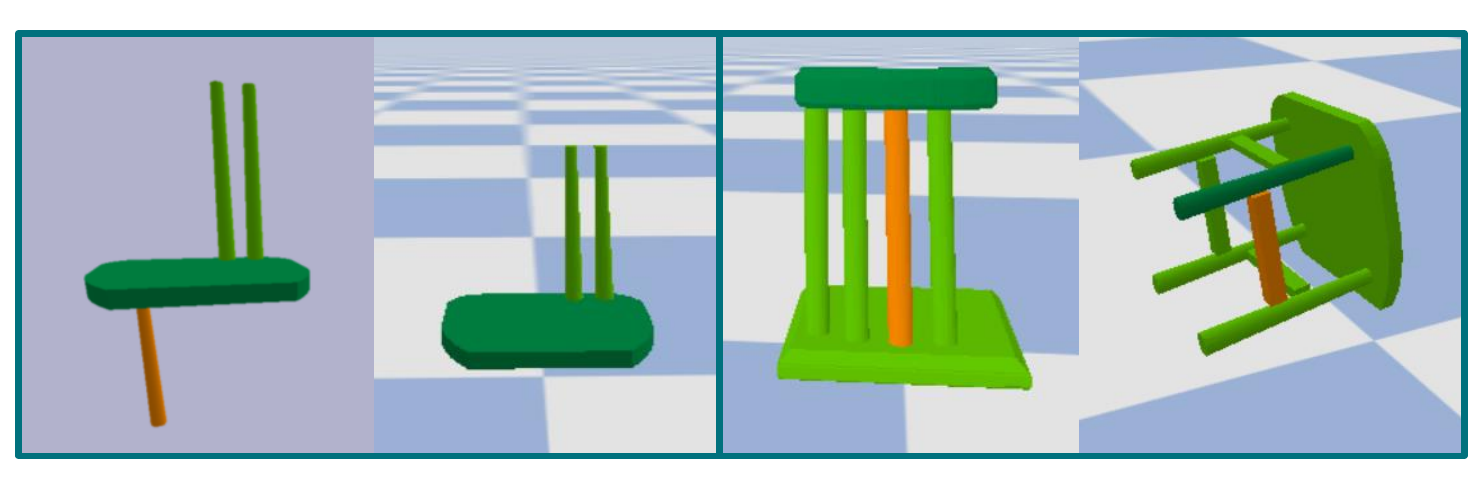}
 \caption{Several types of unsuccessful assembly process. The green parts are motionless, and the orange part is moving towards its target position. 
 Left: moving object (orange) cannot move to the position shown on the left unless colliding with floor or the green ones. 
 Right: moving part (orange) cannot move to the position shown in figure unless colliding with green parts. }
 \label{fig:challenge}
\end{figure}

\subsection{Data Assets}
We select 220 chairs from PartNet, randomly split them into the training set and test set. Based on the assembly difficulty, the training set is further divided into two subsets:~\emph{Easy Train} and~\emph{Hard Train}. Examples for each level can be found in Fig.\ref{fig:result_worobot}.

We make adjustments to these meshes to satisfy collision-free constraints in our task. Leveraging the annotation from PartNet~\cite{mo2019partnet}, we get a fine-grained segmentation for each part. In PartNet, each part's local frame is set to be the center of the chair, rather than each part's own center of mass. We set each part's local frame on its center of mass. We then adjust the relative pose between each part so that there is no penetration between the two parts.

\begin{figure}[t!]
 \centering
 \includegraphics[width=\linewidth]{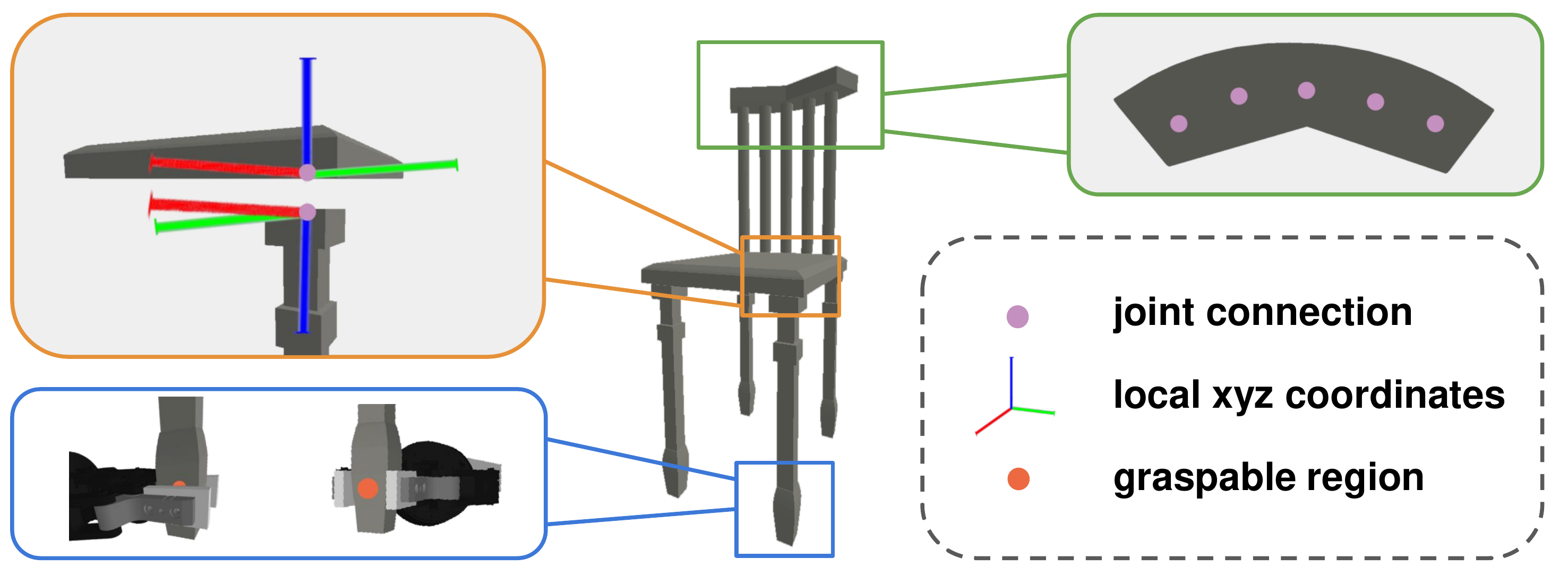}
 \caption{Example of additional annotations over PartNet dataset. 
 Green: Several connection point are annotated on a part. Each joint corresponds to a connection location to a different part.
 Orange: The connection location on two parts is annotated with two points, respectively. For each connection point, we annotate the position and local coordinates.
 Blue: On each part, two graspable regions are annotated. When the gripper is to grasp a part, it can select one of the regions and a direction out of four.}
\label{fig:joint_info}
% \vspace{-4mm}
\end{figure}

Additional annotations include the connection points between two parts and graspable regions, as shown in Fig.\ref{fig:joint_info}. We first calculate the minimum distance between any two parts/meshes of a chair. Once the minimum distance between two parts is detected less than a threshold of 5mm, the two parts are considered as connected. If two parts are treated as connected, we annotate the two points corresponding to the minimum distance between these two parts as the connection points. Then we define a set of normal vector and tangent vector at the contact surface as shown in Fig.\ref{fig:joint_info}. One connection point on a part is depicted by a 9-dim vector: connection point position relative to the part center, its normal vector, and its tangent vector. We also annotate the graspable regions for robots manually. A feasible graspable region must allow collision-free grasp, and each part is annotated with two graspable regions.

%% file: tex/probdef.tex
We formulate the robotic assembly task as a reinforcement learning problem.
In this section, we describe the detailed definitions of states~$\mathcal{S}$, actions~$\mathcal{A}$, rewards~$\mathcal{R}$, and step function $\mathcal{D}$ as follows.

We first introduce the notations used in the paper. Assume a chair $\mathcal{Z}^i$ has $M^i$ parts. Each part in the chair is assigned with a part id~$\mathcal{P}^i_{x}$ where $x \in \{1,2,...,M^i\}$. We adopt the part id assignment order from the PartNet~\cite{mo2019partnet} description. For each part, our simulation provides its geometry description represented as a mesh file. 

Each part $\mathcal{P}^i_x$ has a predefined connection id list $\mathcal{J}^{i}_{x,y}$ where $y \in \{1, 2,...,K^i_x\}$. Here $K^i_x$ is the total number of connection points on the part $\mathcal{P}^i_x$. As shown in the Fig.\ref{fig:joint_info}, for two connected parts, we annotate the corresponding connection point information on each part's mesh file. The connection information on each part contains the contact point 3D position based on object frame and the local xyz-coordinates with the contact point set as the origin denoted as $^\mathcal{J}\mathcal{T}^{i}_{x,y}$. Note that all the 6D poses or transformations in this work are composed of 3D translation and 3D orientation represented as the Euler angles. Besides the connection information, we also annotate the graspable region on each part. These graspable regions are used for robotic grasping. 

At the beginning of each learning episode, we load a chair's full set of parts into our simulation and distribute them randomly on the ground. At each time step, some parts are moved to new poses. To record the part connection status at each time step, we maintain a part connection status tensor $\mathcal{C}^i$ with a shape of $(M^i,M^i,6)$ . If the part $\mathcal{P}^i_u$ and part $\mathcal{P}^i_v$ is not connected, the $C^i(u,v)$ and $C^i(v,u)$ are set to be a six-dimensional zero vector. If the part $\mathcal{P}^i_u$ and part $\mathcal{P}^i_v$ is connected, the $C^i(u,v)$ and $C^i(v,u)$ are set to be the relative transformation between these two parts denoted as $\mathcal{T}^{i}_{t,(u,v)}$ and $\mathcal{T}^{i}_{t,(v,u)}$, respectively. 

\subsection{States}
The state $\mathcal{S}_t$ in our setting is composed of each part's status. The part $\mathcal{P}^i_{x}$ status has two main aspects. One is the geometry description, which includes the shape of the part denoted as $\mathcal{G}^i_x$, the connection information $^\mathcal{J}\mathcal{T}^{i}_{x,y}$  on each part, and the grasping region $^\mathcal{G}\mathcal{T}^{i}_{x}$ on each part. The other is the spatial and connection relationship among these parts. The absolute pose of the part $\mathcal{P}^i_{u}$ at the current time step $t$ denoted as $\mathcal{T}^{i}_{t,u}$ is provided. The relative pose transformation between two parts $\mathcal{P}^i_{u}$ and $\mathcal{P}^i_{v}$ can be directly calculated denoted as $\mathcal{T}^{i}_{t,(u,v)}$ and $\mathcal{T}^{i}_{t,(v,u)}$. The connection status between two parts could be either ``Connected" or ``Not Connected." Here if two parts, $\mathcal{P}^i_{u}$ and $\mathcal{P}^i_{v}$, are recorded to be ``Connected," it indicates that these two parts would be rigidly fixed together and they would share the same rigid 6D transformation. If the connection status between two parts is "Not-Connected", these two parts could be moved independently. The part connection status tensor $\mathcal{C}^i$ is provided. By looping the items in the tensor $\mathcal{C}^i$, the agent could figure out which parts are connected rigidly into a rigid group. Note that the number of groups varies during the assembly process. In the beginning, each part is considered as a rigid group. When the chair is successfully assembled, there is only one rigid group. 

In summary, the states to assemble the chair $\mathcal{Z}^i$ with $M_i$ parts are $(\{\mathcal{G}^i_x, \{^\mathcal{J}\mathcal{T}^i_{x,y}\}_{y=1}^{K^i_x}, ^\mathcal{G}\mathcal{T}^i_{x}, \mathcal{T}^{i}_{t,x}\}_{x=1}^{M^i},\mathcal{C}^i)$

\subsection{Actions}\label{sec:action}
Here we provide two types of assembly settings: 1) the object-centric setting, which does not involve the robots and does not need to consider robotic grasping and robot's motion collision; 2) the full setting, which contains two robots to grasp and connect parts. 

At each time step, we need to select which two parts should be connected through which two connection point. Denote the two selected part ids as $u$ and $v$. The part $\mathcal{P}^i_u$ and $\mathcal{P}^i_v$ has its own connection id list $\{\mathcal{J}^{i}_{u,y}\}_{y=1}^{K^i_u}$ and $\{\mathcal{J}^{i}_{v,y}\}_{y=1}^{K^i_v}$. Denote the two selected connection id as $k$ and $l$. The connection point represented by $\mathcal{J}^i_{u,k}$ on part $\mathcal{P}^i_u$ is connected to the point represented by $\mathcal{J}^i_{v,l}$ on part $\mathcal{P}^i_v$. 
 
Besides the above four parameters~$ (u,v,k,l)$ which are two selected part ids and two selected connection ids, we also need to decide how to move these two $\mathcal{P}^i_u$ and $\mathcal{P}^i_v$ to be connected in the simulation.

In the {\bf object-centric setting}, we would move the $\mathcal{P}^i_u$ from its current pose to the target pose connecting to $\mathcal{P}^i_v$. As illustrated in Fig.\ref{fig:result_worobot}, due to the existence of ground, not all configurations of the group of $\mathcal P^i_v$ allow enough space for $\mathcal P^i_u$ to be connected. Thus, we provide an extra action denoted as $w$ to indicate how to rotate the part~$\mathcal{P}^i_v$. . So the group of $\mathcal P^i_v$ is able to reselect its orientation as action $w$ before path-planning. With a correct selection of $w$, there is a collision-free path to move the part~$\mathcal{P}^i_u$ from its current pose to the relative pose to be merged toward part~$\mathcal{P}^i_v$. Therefore, the action space in the object-centric setup case has five dimensions~$(u,v,k,l,w)$, which are two selected parts, two selected connection ids, the pose id of the second selected part during mating. Once these two parts are attached, the relative pose between them is fixed, and they're free to move as a group under gravity. 

In the {\bf full setting}, after we have selected the two parts~$\mathcal{P}^i_u$ and $\mathcal{P}^i_v$. We need to select the grasp approach direction denoted as $(g_a,g_b)$. All parts can be grasped by the two robots. After grasping a part, the robot could change the part absolute pose in the world frame. Therefore, the action space in the {\bf full setting} has six dimensions~$(u,v,k,l,g_a,g_b)$, which are two selected parts, two selected connection ids, two grasping approach directions. Once these two parts are attached, the relative pose between them is fixed, and they are moved as a group by one robot. The merged group is then placed in a specific region. 

\subsection{Step Function and Rewards}\label{sec:step_fun}
Based on the current state $s_t$ and the selected action $a_t$, the step function describes what is the next state $s_{t+1}$ and what type of reward $r_t$ are returned. If two parts are successfully merged, the agent receives a reward of one point. Otherwise, the agent receives zero reward and the process terminates. If the chair $\mathcal{P}^i$ is fully assembled, the agent receives an extra bonus reward of four points. 
Next, we discuss the detailed processes in our simulation environment. One challenge in the step function is how to deal with the symmetry parts such as the four legs shown in Fig.\ref{fig:teaser}. Note that for those symmetry parts, their local connection annotations and grasping region annotations are all the same. After selecting two parts $\mathcal{P}^i_u$ and $\mathcal{P}^i_v$, we would need to gather all parts which are geometrically equivalent to $\mathcal{P}^i_u$, denoted as $Set(\mathcal{P}^i_u)$. Note that the action at the object-centric setting is $ (u,v,k,l)$ where $(u,v)$ two selected part ids and $(k,l)$ are two selected connection ids.
We verify whether the two parts are able to be connected according to the selected two part ids and connection ids. We search the ground-truth assembled chair to find the part denoted as the $\mathcal{P}^i_{u'}$ which is connected to the part $\mathcal{P}^i_{v}$ through the selected connection id $\mathcal{J}^i_{v,l}$. Accordingly, we find the part denoted
as the $\mathcal{P}^i_{v'}$ which is connected to the part $\mathcal{P}^i_{u'}$ through the selected connection id $\mathcal{J}^i_{u',k}$. If $\mathcal{P}^i_{u'}$ is also in the $Set(\mathcal{P}^i_u)$ and the select part $\mathcal{P}^i_{v'}$ is actually $\mathcal{P}^i_{v}$, these selected two parts and two connection points are reasonable without considering the physical interaction constraints.

In the {\bf object-centric setting}, the selected part $\mathcal{P}^i_v$ would be set to the pose decided by the fifth action $w$. We run the motion-planning module to check whether there is a collision-free path so that the part $\mathcal{P}^i_u$ could be attached to the part $\mathcal{P}^i_v$.

In the {\bf full setting}, denote our two robots as $R_a$ and $R_b$. The remaining two actions $(g_a,g_b)$ represent the two grasping directions for the robots $R_a$ and $R_b$, respectively.  We first use the motion-planning module to query whether there are feasible collision-free paths to grasp the two parts according to the grasping approach directions. If there are no such paths, the step function returns false and a reward of zero. For example, a chair leg connected with a seat cannot be grasped from the seat side because the seat itself has blocked any approach direction to reach the leg. If there are feasible paths, in the simulation, we execute the two robots to grasp the two parts $\mathcal{P}^i_u$ and $\mathcal{P}^i_v$. Thereafter, these two robots would move to two random collision-free positions by moving their mobile bases and maintaining the same configurations of other joints. We use the motion-planning module to find a feasible collision-free path to move the robot $R_a$ to mate the part $\mathcal{P}^i_u$ to the part $\mathcal{P}^i_v$ grasped by the robot $R_b$. If there is no feasible path, the step function returns false and a reward of zero. As the example illustrated in Fig.\ref{fig:result_worobot}, the vertical bar on the back cannot be moved to its target pose unless colliding with either horizontal bar above or the seat. If there are feasible paths, the robots execute the paths in the simulation. Then the robot $R_a$ releases its gripper. We utilize the motion-planning module to find a path to move the new merged part grasped by the robot $R_b$ and place it to one stand as shown in the video. After placing the merged part, the robot $R_a$ and $R_b$ move to new random positions. The step function is then finished, a reward of one is returned.

%% file: tex/approach.tex
In the section, we describe how to train a unified multi-task model to generate actions to assemble different chairs. We first introduce the processes of learning a single-task policy to assemble one chair and then describe how we develop the multi-task model. 
\subsection{Single-task Policy}
Consider the process to assemble a chair $\mathcal{Z}^i$ with the total $M^i$ parts.
At each time step, the state is $(\{\mathcal{G}^i_x, \{^\mathcal{J}\mathcal{T}^i_{x,y}\}_{y=1}^{K^i_x}, ^\mathcal{G}\mathcal{T}^i_{x}, \mathcal{T}^{i}_{t,x}\}_{x=1}^{M^i},\mathcal{C}^i)$. As described in Sec~\ref{sec:action}, the action in the object-centric setup and the full setup are $(u,v,k,l,w)$ and $(u,v,k,l,a,b)$, respectively. In both setups, the actions are discrete. Thus, we adopt the Double-DQN algorithm~(DDQN)~\cite{van2016deep} to learn the policy. 

We first train a model which extracts the geometry feature from the point clouds. The model is an Auto-Encoder based on the PointNet~\cite{qi2017pointnet} and we adopt the Chamfer distance loss~\cite{fan2017point} to optimize the model's weights. In order to train the Auto-Encoder, we create a large set of point clouds sampled from all chairs in PartNet~\cite{mo2019partnet}. We then fix the model's weight and use it as a feature extractor.

We sample a point cloud of each part $\mathcal{P}^i_x$ at its beginning pose $\mathcal{T}^i_{0,x}$ denoted as $X^i_{0,x} \in R^{m \times 3}$ . Then we feed the point cloud $\mathcal{T}^i_{0,x}$ into the above model to extract a geometry feature for each part denoted as $\mathcal{F}^i_{x}$.

After gathering the vector $\{\mathcal{F}^i_{x}\}_{x=1}^{M^i}$, we directly concatenate these vectors with the rest items $\{^\mathcal{G}\mathcal{T}^i_{x}, \mathcal{T}^{i}_{t,x}\}_{x=1}^{M^i}$ and the part connection matrix $\mathcal{C}^i$. Note that all items are reshaped as 1-dim vector before the above concatenation.

Note that different chairs may have different numbers of parts, and different parts may have different numbers of connection points. We set two assumptions: a chair could only have a maximum number of 20 parts; a part could only have a maximum number of 10 connection points. These two assumptions hold within our dataset. We think they are, in general, reasonable. We add zero padding in the concatenated vector if any chair has less than 20 parts or any part has less than 10 connection points. The concatenated vector is then fed into multiple Multilayer Perceptron~(MLP)s to extract a global feature. The feature is decoded to generate the Q values associated with different actions.

\subsection{Multi-task Policy}
We denote these chairs in the training set as $\{\mathcal{Z}_i\}_{i=1}^{N_{train}}$.
After training single-task policies, we now have expert models for each chair $\mathcal{Z}_i$. We then use these expert models to train a multi-task model through imitation learning. Denote Q-function which guides to assemble a chair $\mathcal{Z}_i$ as $Q^{\mathcal{Z}_i}(s,\cdot)$. Note that for discrete action, the $Q^{\mathcal{Z}_i}(s,\cdot)$ is a vector with a size equaling to the action space's dimension. 

We aim to learn a Q-function denoted as $Q(s,t)$ that generates Q values when various chairs are assembled. For each chair $\mathcal{Z}_i$ in the training set, we generate large amounts of training pairs with state $s$ as the inputs and the corresponding Q values $Q^{\mathcal{Z}_i}(s,\cdot)$ as the ground truth annotation. We also apply data-augmentation methods such as adding noise to the point clouds of the data we collected from successful single-task policies.
 
We then adopt two types of loss functions to guide the weight updates. The first loss shown as in Eqn.\ref{eqn:MSE} is a mean square loss between the predicted Q values and the ground-truth Q values. To successfully assemble these parts into a chair, our model at each time step select the action associated with the highest Q value. We adopt a second loss to update the weights such that the ground-truth action index has the highest Q value in our predicted models.
\begin{align}
    &\mathcal{L}_1=\Vert Q(s,\cdot)-\hat{Q}(s,\cdot)\Vert_2\\
    % &\mathcal{L}_2=\left(\frac{Q(s,\argmax_{a} Q_{\mathcal{Z}_i}(s,a))}{\max_a Q(s,a)}-1\right)^2\\
    &\mathcal{L}_2=\max_a Q(s,a) - Q(s,\argmax_{a} Q_{\mathcal{Z}_i}(s,a))\\
    &\mathcal{L}=\mathcal{L}_{1}+\lambda \mathcal{L}_{2}
\end{align}\label{eqn:MSE}
where $\lambda=50.0$

%% file: tex/exp.tex
In this work, we develop a framework for robots to learn to assemble diverse chairs. Our experiments focus on evaluating how effective is our proposed learning-to-assemble approach compared with other baselines.

\begin{figure*}[thb!]
 \centering
 \includegraphics[width=\linewidth]{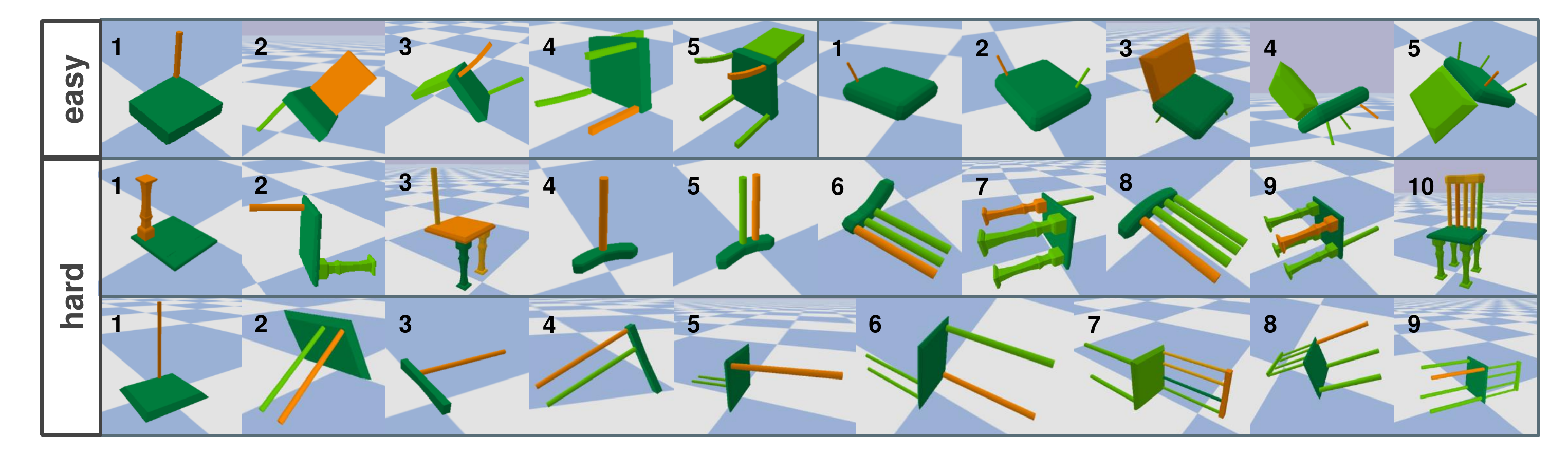}
 \caption{
 Examples of successful object-centric assembly processes. The orange and green parts represent the group of $\mathcal P_u^i$ and $\mathcal P_v^i$, respectively.
 %The dark orange and dark green parts represent the part $\mathcal P_u^i$ and $\mathcal P_v^i$. 
 The time step is increased from right to left, showing how a shape is assembled in order.}
 \label{fig:result_worobot}
\vspace{-5mm}
\end{figure*}

\subsection{Baseline: Motion Planning} 
For the baseline, we adopt a motion planning (MP) algorithm to find the solution path to assemble each chair. The dimension of the state is $6M^i$, where $M^i$ is the total number of the parts from the chair $\mathcal{Z}^i$. Note that we have access to the ground-truth 6D pose of each part when the chair is successfully assembled. These 6D poses together become the goal state in our baseline. 
In the beginning, all parts are positioned at their initial poses as in our {\bf object-centric} setting. We run the RRT-connect~\cite{844730} algorithm to find a collision-free path in the state space, which moves all parts from their initial 6D poses to their goal poses. 
The maximum number of searching operations in the state space is 100k for each chair. 
If no solution is found after checking 100k states in the searching space, the assembly process will be considered a failure. Otherwise, we will get the total planning steps for a successful assembly process. %We assemble each chair for 10 times to mitigate the randomness of planning.

\subsection{Evaluation Metric}
We adopt two different evaluation metrics, which are the success rate and planning steps.

{\bf Success Rate}
        %For both object-centric and full settings, 
        refers to the proportion of successfully assemble a complete chair among all chairs. 
        %For object-centric setting, due to the randomness of motion planning, correct actions are not guaranteed to succeed. There's another evaluation metric of success rate. During evaluation, each chair is tried to assemble 20 times and the times of success is then used to calculate success rate for current model. 
        %For full setting, successful planning path is stored and picked when faced with same state-action pair to shorten planning time consumption. So there's no metric of success rate in this setting.

{\bf Planning Steps} evaluate the total number of attempted states in the searching space during the entire planning process for assembling a chair. It reflects the computation speed of the planning process. The smaller the planning step is, the faster the module is able to find a feasible path.
        % For baseline, it's the number of states attempted in one $6 M^i$ dimension motion planning.
        % For our object-centric and full settings, it's the summation of all attempted states of motion planning in each timestep.

\subsection{Experiment Results and Analysis}
We split the training dataset into the easy training set and the hard training set, according to Sec.~\ref{sec:datasim}. We train one single-task policy for each chair from the training set under the object-centric and the full setting. The maximum training step is set to be 40 thousand for both easy and hard training sets. Then we evaluate the trained model on the corresponding chair it is trained on to report whether it can complete the assembling process. The results are summarized in Table.\ref{tab:result_table}. Under the object-centric setting, our approach achieves a success rate of 66.4\% on the easy training set and 79.4\% on the hard training set. After training single-task policies, we train a multi-task policy under object-centric setup. In the process, we adopt 640 successful single-task policies after data augmentation and test the multi-task policy on 192 unseen chairs. The agent achieves a 74.5\% success rate under the object-centric setting, while baseline can only handle 18.8\% on the same set of chairs. We observe that our multi-task policy can generalize to unseen chairs and outperform the baseline. For successfully assembled chairs, our method is able to plan a series of paths within two minutes on average while baseline requires three times longer. We also check the planning steps for the trained multi-task object-centric model and baseline to assemble an unseen chair. We observe the agent under object-centric setting can assemble a chair within 37.8k steps on average, outperforming baseline's more than 90k steps. This indicates that applying our model to an unseen chair has a higher probability of success and requires much fewer planning steps. 

For the full setting, our approach achieves a success rate of 59.3\% on the training set and 25.8\% on the hard training set. Thereafter, a multi-task policy is trained using 84 successful single-task policies and then tested on 16 unseen chairs. Our approach under full setting achieves a success rate of 50.0\%. Our model has the ability to generalize to unseen chairs. Videos of the assembly processes are available on our project webpage:~\href{https://sites.google.com/view/roboticassembly}{https://sites.google.com/view/roboticassembly}.

\begin{table}[htb!]
\centering
\begin{tabular}{l|ccc}%{@{} lp{0.17\linewidth} |cp{0.19\linewidth} | cp{0.32\linewidth} cp{0.14\linewidth}}
\hline \hline 
Method &  Diff-Level & SuccRate(\%)  & Plan Steps\\
\hline
Ours~(OC) & Easy train & 79.4  & - \\
Ours~(OC) & Hard train & 66.4  & - \\
Ours~(F) &  Easy train & 59.3  & - \\
Ours~(F) & Hard train  & 25.8  & - \\
\hline
Ours~(OC) & Test set   & 74.5  & 37.8k \\
Baseline(OC) & Test set & 18.8  & 91.7k \\
\hline
Ours~(F) & Test set & 50.0  & - \\
\hline
\end{tabular}
\caption{
Comparison of success rates and planning steps for different tasks. 
% Here OC refers to object-centric setting. 
OC: object-centric setting. 
% N refers to full setting.
F: full setting.
% MP refers to our motion planning baseline.
}
\label{tab:result_table}
\vspace{-9mm}
\end{table}

%% file: tex/con.tex
% In this work, 
We develop a robotic assembly simulation environment that supports multi-robots. Based on our simulation environment, we formulate the assembly problem as a concrete reinforcement learning problem and develop a deep reinforcement learning model to successfully assemble a diverse set of chairs. Experiments indicate that when testing with unseen chairs, our approach achieves a success rate of 74.5\% under our object-centric setting and 50\% under our full setting. Our approach outperforms the RRT-Connect baseline by a large margin regarding the success rate and the computation speed.